%% file: isprs_main.tex
\documentclass[a4paper,fleqn]{cas-dc}
\usepackage[authoryear]{natbib}
\usepackage{subcaption}
\usepackage{multirow}
\usepackage{adjustbox}
\usepackage{amsmath,amsfonts,bm}
\usepackage{alphalph}
\usepackage{amssymb}
\usepackage{pifont}
\newcommand{\cmark}{\ding{51}}%
\newcommand{\xmark}{\ding{55}}%

\def\tsc#1{\csdef{#1}{\textsc{\lowercase{#1}}\xspace}}
\tsc{WGM}
\tsc{QE}
\tsc{EP}
\tsc{PMS}
\tsc{BEC}
\tsc{DE}

\input{math_commands.tex}

\begin{document}
\let\WriteBookmarks\relax
\def\floatpagepagefraction{1}
\def\textpagefraction{.001}

\shorttitle{\model}

\shortauthors{Li et~al.}

\title [mode = title]{Cross-View Geolocalization and Disaster Mapping with Street-View and VHR Satellite Imagery: A Case Study
of Hurricane IAN}                      



%
\author[1]{Hao Li}[type=editor,
                        auid=000,
                        bioid=1,
                        orcid=0000-0002-6336-8772]

\cormark[1]
\ead{hao_bgd.li@tum.de}

\author[1,2]{Fabian Deuser}[type=editor,
                        auid=000,
                        bioid=1,
                        orcid=0000-0003-4511-4223]
\ead{fabian.deuser@tum.de}

\author[1,3]{Wenping Yin}[]
\ead{wenping.yin@tum.de}

\author[1]{Xuanshu Luo}[type=editor,
                        auid=000,
                        bioid=1,
                        orcid=0000-0002-6934-5854]
\ead{xuanshu.luo@tum.de}

\author[1]{Paul Walther}[type=editor,
                        auid=000,
                        bioid=1,
                        orcid=0000-0002-5101-5793]
\ead{paul.walther@tum.de}

\author[4]{Gengchen Mai}[type=editor,
                        auid=000,
                        bioid=1,
                        orcid=0000-0002-7818-7309
                        ]
\ead{gengchen.mai@austin.utexas.edu}

\author[5]{Wei Huang}[]
\ead{wei_huang@tongji.edu.cn}

\author[1]{Martin Werner}[type=editor,
                        auid=000,
                        bioid=1,
                        orcid=0000-0002-6951-8022]
\ead{martin.werner@tum.de}

\affiliation[1]{organization={Professorship of Big Geospatial Data Management, School of Engineering and Design, Technical University of Munich},
    city={Munich},
    postcode={85521}, 
    state={Bavaria},
    country={Germany}}

\affiliation[2]{organization={Institute of Distributed Intelligent Systems, University of the Bundeswehr Munich},
    city={Neubiberg},
    postcode={85579}, 
    state={Bavaria},
    country={Germany}}

\affiliation[3]{organization={School of Environment and Spatial Informatics, China University of Mining and Technology},
    city={Xuzhou},
    postcode={221116}, 
    country={China}}
    
\affiliation[4]{organization={Spatially Explicit Artificial Intelligence Lab, Department of Geography and the Environment, University of Texas at Austin},
    city={Austin},
    postcode={78712}, 
    state={Texas},
    country={USA}}
    
\affiliation[5]{organization={College of Surveying and Geo-Informatics,Tongji University},
    city={Shanghai},
    postcode={200092}, 
    country={China}}

\cortext[cor1]{Corresponding author}



\input{abstract.tex}




\maketitle

\input{intro.tex}

\input{related_work}

\input{method}

\input{experiment}

\input{discussion}

\input{conclusion}

\section*{Acknowledgements} The authors gratefully acknowledge the computing time granted by the Institute for Distributed Intelligent Systems and provided on the GPU cluster Monacum One at the University of the Bundeswehr Munich.

\section*{Disclosure statement} No potential conflict of interest was reported by the author.


\printcredits

\bibliographystyle{cas-model2-names}

\bibliography{reference}



\end{document}

%% file: math_commands.tex
\def\eqref#1{equation~\ref{#1}}

\def\1{\bm{1}}

\def\rmI{{\mathbf{I}}}

\def\vtheta{{\bm{\theta}}}

\def\sR{{\mathbb{R}}}

\newcommand{\model}{CVDisaster}

\newcommand{\geolocmodel}{\model-Geoloc}
\newcommand{\imgclsmodel}{\model-Est}

\newcommand{\dataset}{CVIAN}

%% file: abstract.tex
\begin{abstract}
Nature disasters play a key role in shaping human-urban infrastructure interactions, Effective and efficient response to natural disasters is essential for building resilience and sustainable urban environment. Two types of information are usually the most necessary and difficult to gather in disaster response. The first information is about the disaster damage perception, which shows how badly people think that urban infrastructure has been damaged. The second information is geolocation awareness, which means how people's whereabouts are made available. In this paper, we proposed a novel disaster mapping framework, namely \model{}, aiming at simultaneously addressing geolocalization and damage perception estimation using cross-view Street-View Imagery (SVI) and Very High-Resolution satellite imagery. \model{} consists of two cross-view models, where  \geolocmodel{} refers to a cross-view geolocalization model based on a contrastive learning objective with a Siamese ConvNeXt image encoder and \imgclsmodel{} is a cross-view classification model based on a Couple Global Context Vision Transformer (CGCViT). Taking Hurrican IAN as a case study, we evaluate the \model{} framework by creating a novel cross-view dataset (\dataset) and conducting extensive experiments. As a result, We show that \model{} can achieve highly competitive performance (over 80\% for geolocalization and 75\% for damage perception estimation) with even limited fine-tuning efforts, which largely motivates future cross-view models and applications within a broader GeoAI research community. The data and code are publicly available at: https://github.com/tum-bgd/CVDisaster.

\end{abstract}

%% file: intro.tex
\section{Introduction}

Given the fast development in Remote Sensing (RS) technology, the availability of large-scale and high-quality Earth observation (EO) data has significantly benefited timely humanitarian responses to natural disasters \citep{van2000remote, dong2013comprehensive, 2023_GWME_Hao}. Meanwhile, recently, Street View imagery (SVI) has gained significant momentum in urban studies and computer vision in the last few years \citep{zhang2018measuring,zhang2019social,biljecki2021street}, and has shown great potential in complementing traditional satellite imagery analysis by providing a unique and informative cross-view perspective on the ground \citep{zhu2022transgeo}.

In a disaster mapping scenario, two types of information are critical for timely and accurate disaster response and relief. The first type of information is the disaster damage perception, which refers to the ways in which individuals and groups evaluate, subjectivize, and perceive damages to the urban built environment due to the disaster. This information is usually estimated from RS data based on expert knowledge and intensive manual efforts. The second type of information is geolocation awareness, which is basically how accurately people can geographically locate themselves on the map. By combining both information, an ideal disaster mapping framework is able to simultaneously estimate human perception of the damage levels and provide accurate geolocations in the affected areas. 

However, it is not a trivial task to build such a framework due to two major challenges: on the one hand, traditional RS data can become insufficient for fine-grained damage perceptions, especially for distinct and sophisticated urban contexts, where a potential solution is to combine satellite imagery with the emerging source of SVIs to ensure a more fine-grained and cross-view of urban disaster damage perception. On the other hand, existing geolocalization approaches are often not satisfying, because they predominantly depend on satellite navigation systems, such as GPS, Galileo, and BeiDou, which typically lack the appropriate accuracy required for disaster response. Meanwhile, urban context and weather conditions can bring another dimension of complexity where satellite signals are blocked. Fortunately, we have enough ingredients to address the latter challenge as cross-view geolocalization with satellite and street-view imagery offers a sensible alternative. Herein, this technique can match real-time SVI obtained from carriers against a collection of satellite imagery with known geolocations so that the geographical coordinates of SVI can be decided. To the best of our knowledge, there is no such disaster mapping framework exists that can achieve damage perception and cross-view geolocalization at the same time. 

\begin{figure*}[!htbp]
\centering
\centering
\includegraphics[width=\linewidth]{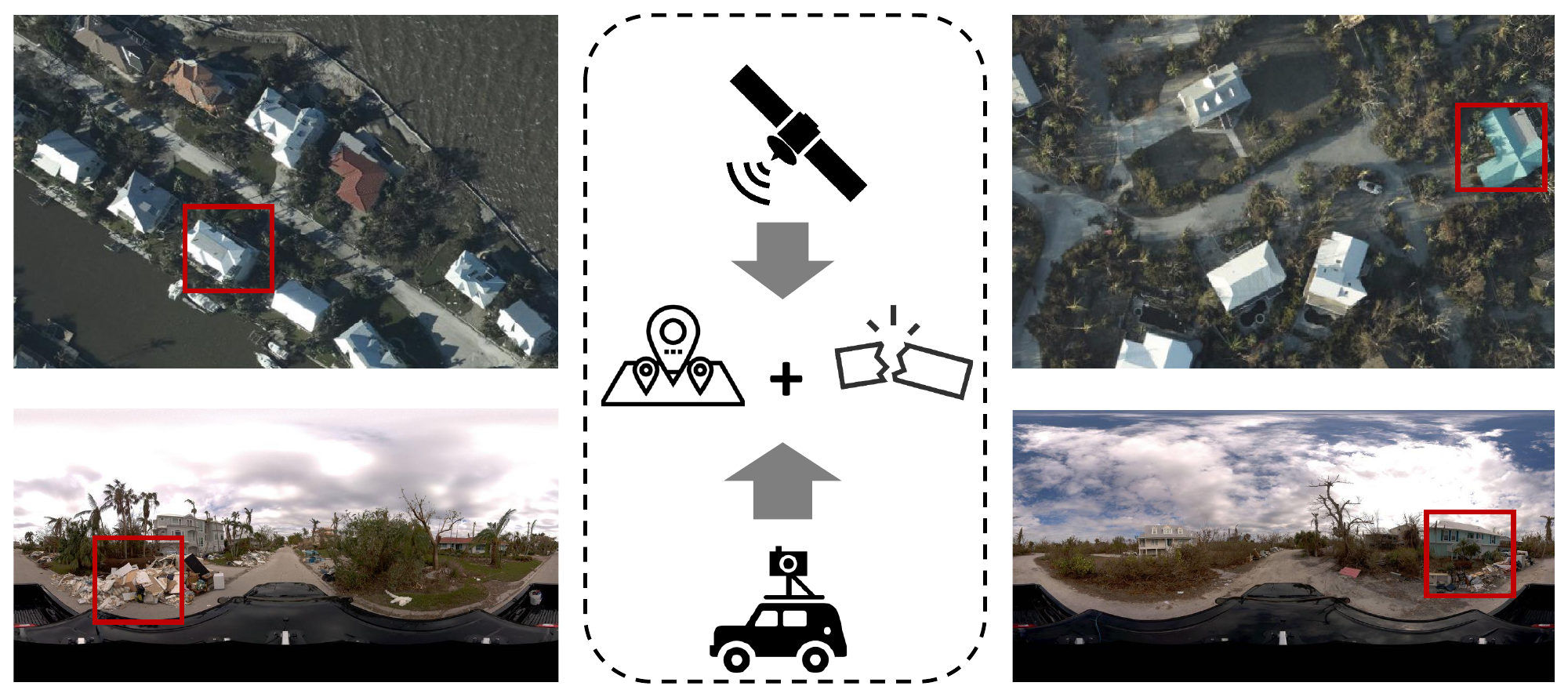}
\caption{An overview of the proposed framework for \textbf{Cross-view} Geolocalization and \textbf{Disaster} mapping with street-view and satellite imagery, namely \textbf{\model{}}.}
\label{fig:overview}
\end{figure*}

In this paper, we fill the aforementioned research gap by developing a novel disaster mapping framework - \textbf{\model{}} - (see Figure \ref{fig:overview}). Specifically, our framework addresses the damage perception estimation and cross-view geolocalization at the same time by leveraging the head-view satellite and street-view imagery using state-of-the-art Geospatial Artificial Intelligence (GeoAI) models. To validate the proposed framework, we conducted a case study in Sanibel Island, Florida, which was hit by Hurricane Ian in 2022. Intensive experiments show the great potential of \textbf{\model{}} in providing timely damage perception and geolocation awareness with competitive accuracy, leading to substantial advantages for future disaster response applications. Moreover, we made the case study dataset (i.e., \dataset{}) openly available to encourage related research in both computer vision and disaster mapping communities. 

In Section \ref{related_work}, we give an overview of related works regarding state-of-the-art disaster mapping, street-view image-based urban analysis, and cross-view geolocalization, respectively. In Section \ref{method}, we elaborate on the detailed methodology design of the proposed framework, ranging from the problem statement to the training and inferencing of both geolocalization and damage perception estimation models. Next, Section \ref{experiment} shows the experimental results from the case study of Hurricane IAN and summarizes the key findings, followed by Section \ref{dicussion} presenting a critical reflection of limitations and identifying future works. Last but not least, Section \ref{conclusion} concludes the paper by highlighting the scientific contributions to a broader community.

%% file: related_work.tex
\section{Related Work} \label{related_work}

\subsection{GeoAI for Disaster Mapping and Localization}

Disaster mapping refers to the capability for even non-profession to assist in disaster response situations via mapping and other spatial analysis \citep{herfort2021evolution, li2022improving}. The concept of disaster mapping has been successfully used to support disaster response and humanitarian aid activities, especially under a disaster scenario, where successful examples include the mapping tasks during the 2017 Hurrican Harvey \citep{feng2020flood}, the 2019 Cyclone Idai and Kenneth in Mozambique \citep{li2020exploration}, and the 2023 Turkey Syria Earthquake \citep{2023_Turkey_Earthquakes}. However, considering the time-crucial nature of disaster responses and humanitarian aid, traditional disaster mapping workflows become less efficient and unsatisfactory in covering a large-scale area and providing timely damage assessment within a rather short time. In this context, the emergence of high-resolution satellite imagery allows for faster and better disaster mapping with GeoAI techniques \citep{salcedo2020machine, werner2022atlashdf}, thus providing a promising solution to address this challenge that local stakeholders currently encounter. Early works in this direction \citep{Herfort2019, huck2021centaur} report an interesting finding on improving the speed and accuracy of disaster mappings via a machine-assisted manner. In the meantime, there is a stream of GeoAI research focusing on extracting accurate location information during disasters, mainly from social media text data (e.g., Twitter) \citep{kumar2019location, hu2020people, mihunov2020use, hu2022gazpne, hu2023geo}. One famous example is the news article published in the U.S. National Public Radio, titled “Facebook, Twitter Replace 911 Calls For Stranded In Houston”, which reported how affected people by Hurricane Harvey in 2017 used social media to share their location and asked for help, which significantly helps the rescue team to locate and reach those people in need. One can find a comprehensive survey on location reference recognition in \cite{hu2023survey}.

However, a majority of existing disaster mapping and localization approaches either rely on post-disaster satellite imagery analysis for damage assessment or use geoparsing tools to georeference a social media text. Therefore, there is a pressing need for an intelligent disaster mapping and geolocalization solution, ideally within a single framework. To the best of our knowledge, \textbf{\model{}} is the first such integrated framework that can achieve large-scale damage perception and cross-view geolocalization at the same time.

\subsection{Street-view Imagery for Urban Analytics}

Due to its emerging availability, SVI has become a crucial data source for urban studies. ~\cite{diakakis2017identifying} conducted a comprehensive review of the applications of SVI in urban research, highlighting its growing significance in urban analysis. Their study indicates that most urban research utilizing SVI relies on Google SVI (GSVI). However, crowdsourced platforms like Mapillary and KartaView are also rapidly evolving and becoming key tools in urban research.

In urban analysis, SVI is extensively applied across various fields, such as the maintenance of spatial data infrastructure, studies of urban morphology and perception, and traffic flow analysis. For instance, ~\cite{kim2020citycraft} and ~\cite{li2023semi} inferred urban features based on SVI to generate 3D urban models. ~\cite{krylov2018automatic} effectively detected utility poles and traffic signals using GSVI, demonstrating the unique efficacy of SVI in identifying streetlights and traffic signs. In urban morphology analysis, many scholars have estimated urban geometric indicators using SVI to study microclimates and light pollution. ~\cite{hu2020investigation} and ~\cite{cicchino2020not} extracted road variables from SVI to analyze the safety of walking and cycling in urban areas.

Researchers also used SVI to extract information about human health and well-being. By matching participants' movement trajectories with SVI, one can analyze the environmental features residents encounter in their daily activities, providing robust data support for public health policy-making. For example, ~\cite{nguyen2018neighbourhood} investigated GSVI to extract derived indicators such as street greenery, crosswalks, and building types to describe the built environment at the postal code level in three US cities. The study found a correlation between community characteristics and the prevalence of obesity and diabetes. In addition to these key indicators, ~\cite{keralis2020health} demonstrated that factors such as overhead visible wires and whether roads are single-lane are associated with various health outcomes, including diabetes, psychological distress, and alcohol consumption. Further related studies include analyzing residents' air pollution exposure, stress levels, and infectious diseases based on street-view data \citep{apte2017high, han2022measuring, psyllidis2023cities}.

More importantly, SVI plays an increasing role in disaster response, particularly in long-term recovery and reconstruction planning. It helps decision-makers understand changes in disaster-affected areas, providing crucial references for future disaster prevention and urban planning. ~\cite{curtis2012spatial} and ~\cite{curtis2013using} explored recovery after tornadoes, hurricanes, and wildfires using GSVI. ~\cite{mabon2016charting} utilized GSVI from the evacuation zone around the Fukushima Daiichi Nuclear Power Plant to assess dynamic disaster recovery methods. Additionally, SVI has been used in disaster emergency response and risk assessment. ~\cite{diakakis2017identifying} used GSVI to identify the probability of buildings in Athens being flooded. ~\cite{naik2016flooded} designed a crowdsourced sensing system for disaster response during catastrophic flooding in Chennai, India, helping residents in flood-affected areas and reducing casualties. SVI provides detailed ground-level information, such as the condition of damaged buildings, the extent of street flooding, and the state of infrastructure. This information is crucial for disaster assessment and emergency response. By combining SVI with RS data, we can obtain more accurate and comprehensive disaster information, thereby enhancing the precision and efficiency of disaster response and supporting post-disaster recovery and reconstruction. However, research that integrates SVI with RSdata in a disaster response scenario is still limited.

\subsection{Cross-view Geolocalization}

Unlike the single-image geolocalization task \citep{weyand2016planet,cepeda2023geoclip,zhou2024img2loc}, cross-view geo-localisation enhances classic location-based services and navigation systems by matching ground-level imagery with overhead imagery. This enables accurate positioning in GNSS-denied environments, e.g., during a disaster. ~\citet{workman2015wide} showed the superiority of CNN-based features for localizing a wide-ranging dataset with crawled Flickr images across the USA. In subsequent work, they introduced the first cross-view geo-localisation dataset, namely CVUSA~\cite{zhai2017predicting}. This dataset leverages street-view images from GSVI all across the US to match them against overhead imagery to locate the street-views. Since then multiple datasets have arisen with different focuses. CVACT~\cite{liu2019lending} aimed for a larger test set than CVUSA and included the region of Canberra, Australia, to test for cross-domain generalization. As an alternative to ground-level imagery, University-1652~\cite{zheng2020university} introduced drone views of buildings to match them against overhead imagery. Unlike CVUSA and CVACT, which rely on center-aligned street-view images for matching to satellite imagery, VIGOR~\cite{zhu2021vigor} uses a novel approach. This method allows multiple street view images to be matched to a single satellite image at different positions, allowing precise regression of the exact offset. None of the previously released datasets have specifically addressed cross-view geo-localization in disaster scenarios, which involve unique challenges such as destructed and altered environments.

By exploiting image similarities and differences, cross-view geolocation is characterized by contrastive learning. \citet{vo2016localizing} pioneered soft-margin triplet loss and set a long-standing loss standard for this task. Further work introduced specialized aggregation methods like the NetVLAD layer~\cite{Hu_2018_CVPR} or the SAFA-module~\cite{shi2019spatial}, enhancing the ability to capture and aggregate discriminative features from cross-view images. \citet{zhu2022transgeo} are the first to introduce the Transformer architecture in this domain and following work by ~\citet{zhu2023simple}, they utilized the MLP-Mixer architecture with further performance gains. ~\citet{deuser2023sample4geo} introduced hard negative sampling based on the geographical distance as well as feature similarity and showed superior performance. ~\citet{fervers2023c} enhanced this previous work with a second stage for re-ranking the results and improved overall retrieval performance.

%% file: method.tex
\section{Methdology} \label{method}

\subsection{Task statement}

\begin{figure*}[!htbp]
\centering
\centering
\includegraphics[width=\linewidth]{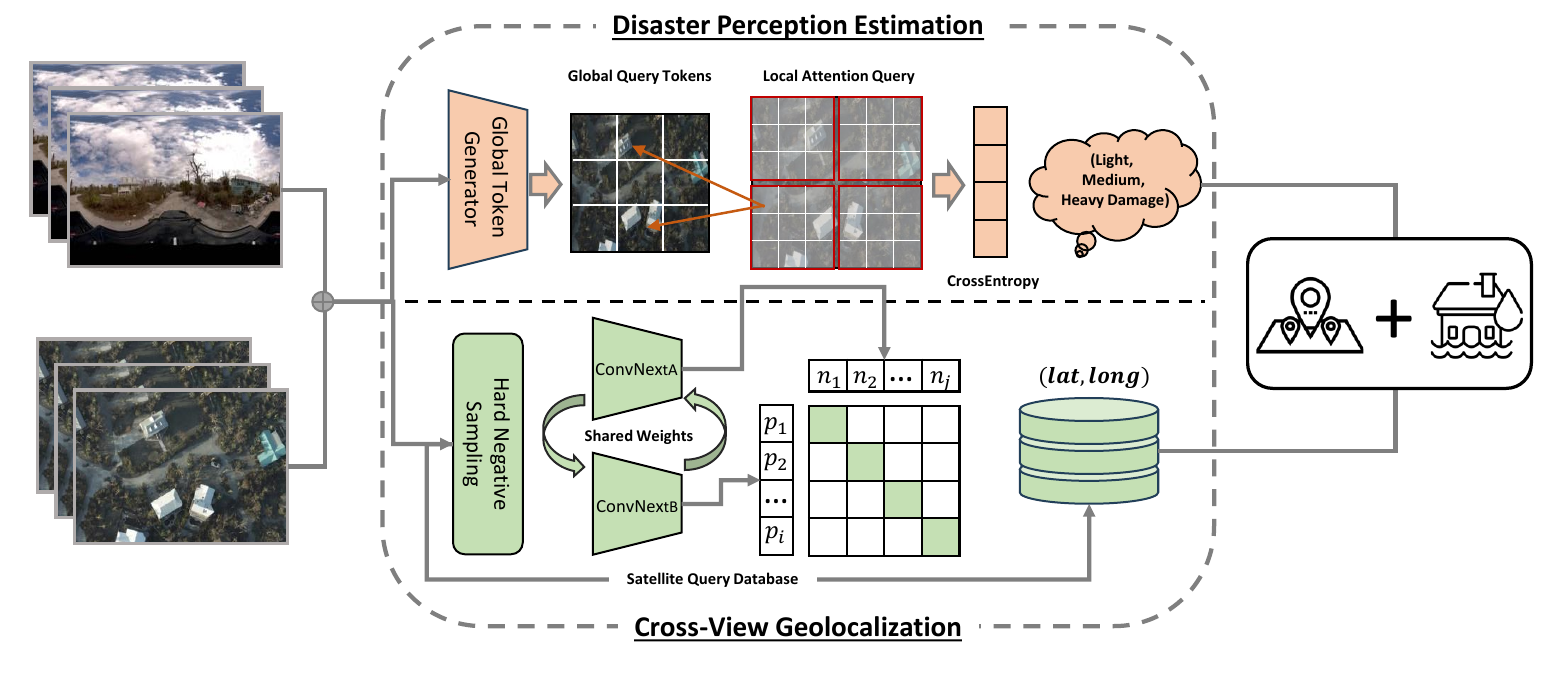}
\caption{The proposed framework, namely \model{}, addresses two key tasks simultaneously, which are 1) \geolocmodel{}: cross-view disaster perception estimation using coupled Global Context Vision Transformer; 2) \imgclsmodel{}: cross-view geolocalization via contrastive learning.}
\label{fig:task_statement}
\end{figure*}

Given a set of street-view imagery $\{ L_s \}$ and satellite imagery $\{ L_a \}$ with $G_a$ refers to the geographical locations (e.g., longitude and latitude) of satellite imagery, our objective to learn a cross-view embedding space $\sR_{CV}$ (e.g., via a non-linear function $f(L_s, L_a) \rightarrow \sR_{CV}$) in which two tasks are solved simultaneously: 1) each street-view imagery $L_s$ is close to its corresponding satellite imagery $L_a$ in the embedding space $\sR_{CV}$ so that the correct geographical location can be retrieved based on their similarities in the embedding space; 2) each pair of street-view and satellite imagery $\{ L_s, L_a \}$ is close to all other pairs where a similar level of damage perception are observed. Figure \ref{fig:task_statement} shows how we achieve this objective by integrating two GeoAI models (i.e., \geolocmodel{} and \imgclsmodel{}), namely the disaster perception estimation model and the cross-view geolocalization model, into a single framework \model{}. In the rest of the section, we will elaborate on the detailed design specifics and model choice.

\subsection{Cross-view Geolocalization via Contrastive Learning}

In this paper, we formulate \geolocmodel{}, the task of cross-view geolocalization, as a imagery retrieval problem, where an image encoder $f()$ is a 
nonlinear function $f(\rmI_i,\vtheta):\sR^{H\times W\times 3} \rightarrow \sR^D$, which is parameterized by $\vtheta$ and maps the input image feature space (i.e., spatial dimension of $H\times W$ with three RGB bands) into a vector embedding representation of $D$ dimension. Herein, cross-view means that $\rmI_i = \{ L^i_s,L^i_a \}$ consisting of paired colocated SVI and satellite imagery, so that the corresponding geo-coordinates $G_a$ from satellite imagery can be queried 
to use as the geographical coordinates of the input SVI. In this setting, two factors are of key importance for a good cross-view geolocalization model, which are the used image encoder $f()$ and the vector embedding representation in the learned feature space $\sR^D$.

\subsubsection{Siamese Image Encoder with the modern ConvNeXt}

To build a rock-solid image encoder for both SVI and satellite imagery, we follow the design in \cite{deuser2023sample4geo} by using a Siamese network that uses the modernized ConvNeXt as a backbone \citep{liu2022convnet}. Similar to the classic ResNet\citep{he2016deep}, ConvNeXt belongs to the Convolution Neural Network (CNN) family, which follows the classic sliding-window, fully convolutional paradigm, but brings in a list of modern neural architecture designs specificity for performance boosting, especially for 
high-resolution input, such as satellite imagery. 

\begin{figure*}[!htbp]
\centering
\centering
\includegraphics[width=\linewidth]{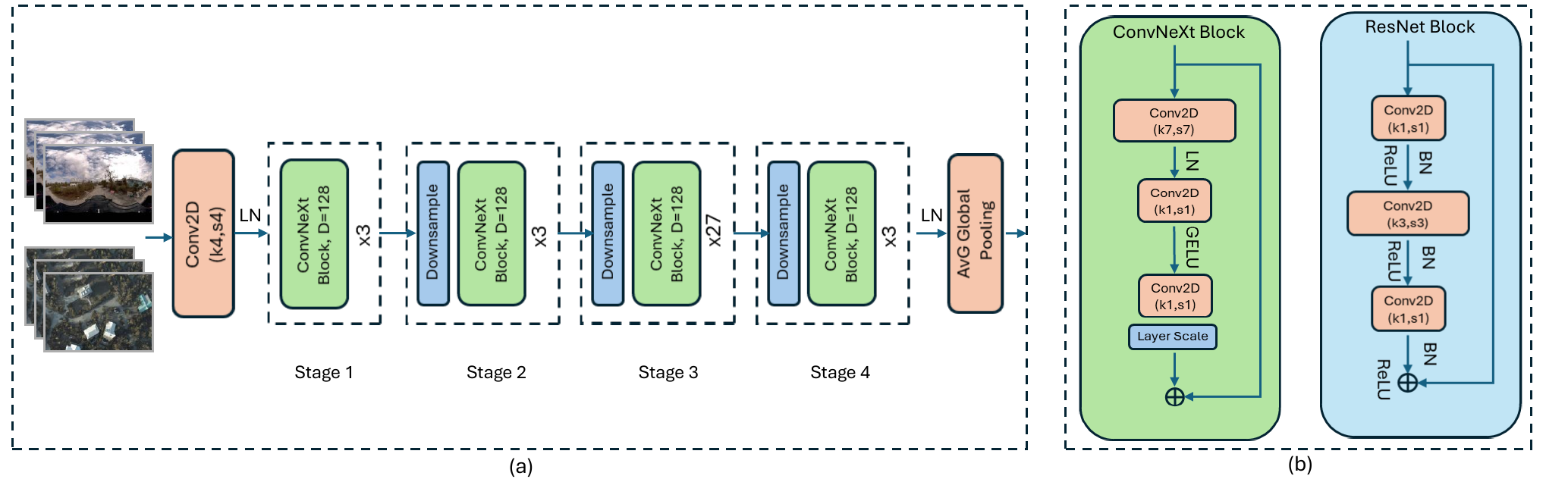}
\caption{The siamese image encoder for cross-view geolocalization using (a) a four-stage ConvNeXt; (b) the comparison of ConvNeXt and ReseNet blocks. }
\label{fig:ConvNeXt}
\end{figure*}

The key motivation for using ConvNeXt as the image encoder $f()$ is actually intuitive: first, it keeps the simplicity and effectiveness of classic CNN then modernizes the ResNet step by step towards the modern Swin Transformer \citep{liu2021swin} style to ensure performance gain. Figure \ref{fig:ConvNeXt} shows the architecture of the 4-stage ConvNeXt network and highlights a comparison between ConvNeXt and ResNet blocks. Herein, it is necessary to notice the following modification w.r.t a classic ResNet model.

\textbf{Stage Compute Ratio:} For classic ResNet, the computation distribution across different stages are decided empirically. For example, ResNet50 is featured with a number of blocks distributed into four stages with a ratio of (3,4,6,3), which makes the convolution operation heavy already in an early stage. One change in Swin Transformer is to reduce the stage compute ratio to 1:1:9:1, which has been introduced to ConvNeXt as well. As a result, the number of blocks in ConvNeXt50 becomes (3,3,9,3).

\textbf{Patchify Layer:} As natural images are inherently redundant, a common practice in the classic ResNet family is to use a stem cell for aggressively down-sampling. However, ViT's patch encoder makes this even more aggressive by adopting a large kernel size and non-overlapping convolution, namely the "patchify" layer. Similar designs are adopted in the new ConvNeXt with a $4\times4$ non-overlapped convolution layer to accommodate the network's multi-stage nature.

\textbf{Inverted Boottleneck and Large Kerner:}
Following a similar idea in the Transformer block, the ConvNeXt block also uses an inverted bottleneck by keeping the dimension of the hidden layer four times of the input dimension. This idea has been proven to be beneficial in the popular MobileNetV2 \citep{sandler2018mobilenetv2} and many more advanced CNN models \citep{koonce2021efficientnet}. Moreover, the ConvNeXt benefits from its larger kernel-sized convolution design, which brings a significantly better performance based on the \cite{liu2022convnet}. As a prerequisite for a larger kernel, the depthwise convolution layer is placed prior to the dense convolutional layers as shown in the comparison of Figure\ref{fig:ConvNeXt}. 

\textbf{Micro-scale Modification:} The modification involves a list of micro-scale improvements, mostly related to the activation function and normalization layer. For instance, the ReLU used in ResNet is replaced by Gaussian Error Linear Unit (GELU) \citep{hendrycks2016gaussian}, which is in fact a smoother variant of ReLU commonly used in modern transformer models. The Batch Normalization (BN) is replaced by the simpler Layer Normalization \citep{ba2016layer} (LN). Furthermore, the downsampling layers are added only between two different stages which are also inspired by the design of Swin Transformers.

Based on this modernized ConvNeXt backbone, we build a siamese network (Figure \ref{fig:ConvNeXt}) as our image encoder $f()$ for both SVI and satellite imagery by adapting the network input to different spatial dimensions. Noticeably, although the Siamese network is trained on cross-view imagery, the inference can handle a single input of SVI as a query base.

\subsubsection{Contrastive Learning with Hard Negative Sampling }

The key to cross-view geolocalization is how to train a siamese ConvNeXt so that one can obtain the desired vector embedding representation of $\rmI_i = \{ L^i_s,L^i_a \}$ in the learned feature space $\sR^D$. Herein, we considered two factors to ensure efficient and effective representation learning in this cross-view setup: 1) contrastive pre-training on large-scale datasets and 2) fine-tuning with new cross-view imagery from the case study area. 

Given the popularity of cross-view geolocalization, there are mainly three large-scale datasets, namely CVUSA \citep{workman2015wide}, CVACT \citep{liu2019lending}, and VIGOR \citep{zhu2021vigor}, which have been made available to the research community. Different in their data sizes, landscape, and sample density, these three datasets form a good basis for pre-training a cross-view geolocalization model to gain nice general-sense vector representations. In this context, we pre-trained the siameses ConvNeXt network on all three datasets (i.e., CVUSA, CVACT, and VIGOR) by using the contrastive learning objective.

Following the "cluster" hypothesis that "closely associated documents tend to be relevant to the same requests" \citep{Ellen1985}, the most common approach of contrastive learning is to simultaneously minimize the distance between the embeddings of the anchor $t_a$ and the positive image $t_p$ while maximizing the distance to the negative sample $t_n$. Therefore, a simple Triplet loss function looks like the following:

\begin{equation}
    L_{triplet} = [ ||f(t_a)-f(t_p)||_2 - ||f(t_a)-f(t_n)||_2 +a]_{+}
\end{equation}

Here, $f()$ is the aforementioned image encoder (e.g., ConvNeXt) whose parameter $\vtheta$ will be learned. To prevent the encoder from pushing the negative image without limitation, a rectifier term with margin $m$ is introduced to keep the maximum distance between the anchor and negative smaller than $m$.

\begin{figure*}[!htbp]
\centering
\centering
\includegraphics[width=\linewidth]{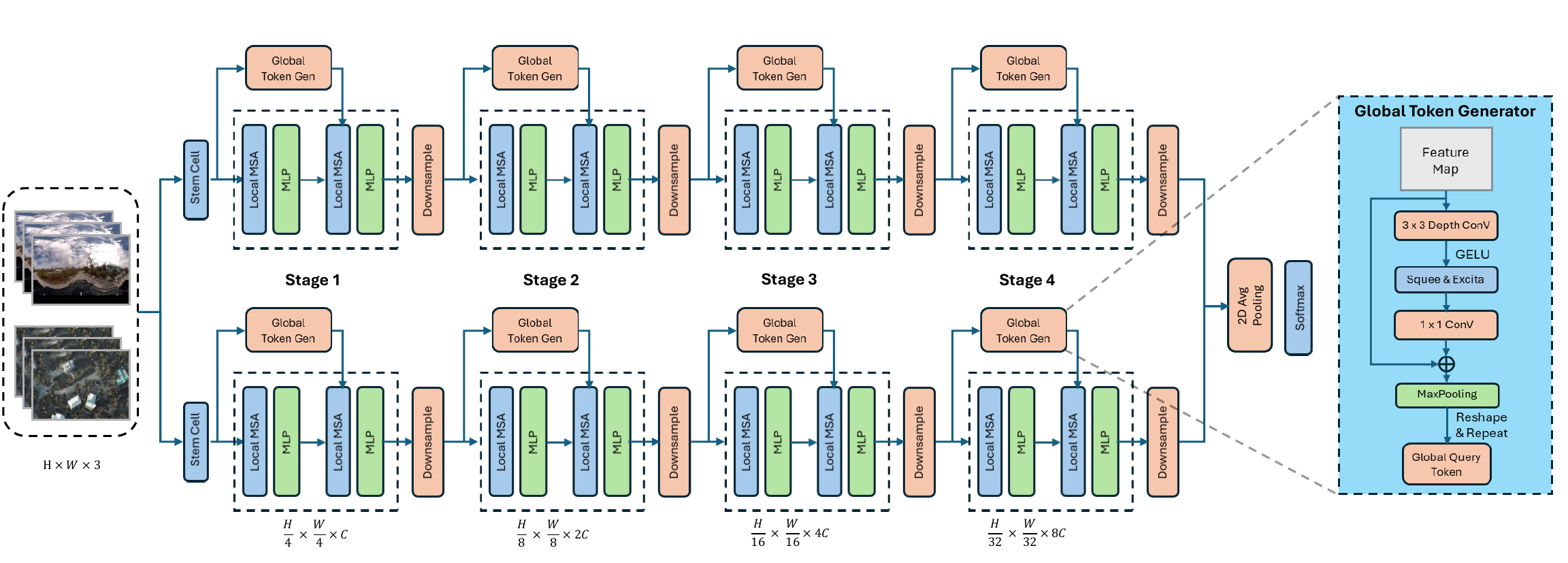}
\caption{Coupled Global Context Vision Transformer for Cross-View Imagery Classification and Damage Perception Estimation. The global token generator is highlighed.}
\label{fig:CGCViT}
\end{figure*}

Compared to the triplet loss, the InfoNCE \citep{oord2018representation,radford2021learning} loss is often considered more robust as it is able to make use of all available negative samples with the batch. Specifically the InforNCE, in a supervised learning setting, computes categorical cross-entropy loss to identify the positive sample amongst a set of negative samples \citep{weng2021contrastive}. Given a context vector $c$, the positive sample is drawn from a conditional distribution $p(x|c)$, where $(N-1)$ negative samples are drawn from the same distribution $p(x)$ but without condition. In this context, the probability of correctly selecting the positive samples can be formulated as follows:

\begin{equation}
    p(C=\texttt{pos} \vert X, \mathbf{c}) = \frac{f(\mathbf{x}_\texttt{pos}, \mathbf{c})}{ f(\mathbf{x}_\texttt{pos}, \mathbf{c}) + \sum_{j=1}^{N-1} f(\mathbf{x}_j, \mathbf{c}) }
\end{equation}

Here, N is the total number of samples in a batch, and $f(\mathbf{x}, \mathbf{c}) \propto \frac{p(\mathbf{x}\vert\mathbf{c})}{p(\mathbf{x})}$ is the similarity or scoring function between two samples. 

Then, the InforNCE loss tries to optimize the negative log probability of correcting selecting the positive samples, thus can be calculated as follows: 

\begin{equation}
    \mathcal{L}_\text{InfoNCE} 
    = - \mathbb{E} \Big[\log p(C=\texttt{pos} \vert X, \mathbf{c}) \Big]
\end{equation}

Although InforNCE has been intensively used in unsupervised and self-supervised representation learning \citep{mai2023csp,vivanco2024geoclip,guo2024spatialscene2vec}, it also offers a promising way for supervised representation learning in this cross-view setup. In this paper, we leverage the InforNCE as our contrastive learning loss in both the pre-training and fine-tuning stages for cross-view geolocalization. During the fine-tuning, we take the model weights pre-trained on CVUSA data given its relatively large size and geographical closeness, then fine-tune the model on the new cross-view imagery collected from the study area in Sanibel Island (Florida, USA) after the Hurricane IAN. To this end, we also compare the geolocalization performance with and without the fine-tuning stage in Section \ref{experiment} as an ablation study.

\subsection{Damage Perception Estimation with Cross-view Imagery}

Herein, \imgclsmodel{}, specifically the task of damage perception estimation, is tackled as a multi-class image classification problem. 
Similarly to the geolocalization task, we define an image encoder $f()$ as a 
nonlinear function $f(\rmI_i,\vtheta):\sR^{H\times W\times 3} \rightarrow \sR^B$ parameterized by $\vtheta$ and would map the input image feature space (again spatial dimension of $H\times W$ with RGB three bands) into a vector embedding representation of $B$ dimension, but following by a softmax classification layer. In this manner, we can use exactly the same cross-view imagery pairs (e.g., $\rmI_i = \{ L^i_s,L^i_a \}$) to simultaneously estimate the damage perception level of the place when another model is trying to decide where SVI are collected. 

Although this is a straightforward model design, 
we argue that this can bring key advantages for \model{} against existing disaster mapping approaches \citep{li2020exploration, Herfort2019, hu2023geo}. On the one hand, the data preprocessing is synchronized with zero overhead for preparing two datasets for distinct tasks (i.e., geolocalization and disaster mapping). On the other hand, the cross-view imagery can provide a unique combination of observation angles and opportunities to fasten and automate the traditional post-disaster survey with inherent geolocation metadata immediately available during the survey. This can be extremely helpful in such a time-crucial application scenario.

In the rest of this section, we will elaborate on how we tackle the damage perception estimation task in \model{} using the modern GeoAI-based imagery classification approach, specifically the couple GCViT model.

\subsubsection{Coupled Global Context Vision Transformer}

To tackle this cross-view image classification task, we develop a coupled GCViT model (CGCViT) as depicted in Figure \ref{fig:CGCViT}) including two separate branches for SVI and satellite imagery, respectively. Unlike the siamese ConvNeXt, the design of CGCViT is driven by two special considerations: first, the appearance of disaster damages from two perspectives (head-view and street-view) differs significantly, therefore, requires highly-distinct image encoders $f()$ or sets of parameter $\vtheta$; second, CGCViT can benefit from the complementary prediction capabilities learned from cross-view pairs at the same time. Moreover, the inference process also differs as the classification of damage perception level always relies on both views while the geolocalization inference actually uses only SVI imagery to query an existing satellite database. This is also why there is a single weight-shared image encoder designed for the cross-view geolocalization task.

As a backbone network, the core idea of GCVit is to advocate short- and long-range spatial dependencies with a multi-resolution architecture where self-attention is still computed in local windows but can reach long-range patch via global tokens \citep{hatamizadeh2023global}. Given a cross-view imagery pair $\rmI_i = \{ L^i_s,L^i_a \}$ with the same dimension of $\sR^{H\times W\times 3}$, the CGCViT consists of two branches of GCViT following by a 2D average pooling layer and a softmax classifier. Each branch will include four stages of local and global self-attention modules similar to ConvNeXt\citep{liu2022convnet} and Swin Transformer \citep{liu2021swin}, but with an increasing number of channels and decreasing spatial resolutions, both by a factor of 2. Herein, the difference between local and global self-attention modules lies in the access to global queried features from the global query generator. 

\textbf{Global Token Generator:} As highlighted in Figure \ref{fig:CGCViT}, the key advantage of GCViT comes from the fact that global attention is able to query long-range perception fields while keeping the local attention window unchanged. Herein, the global query token or so-called global self-attention can be pre-computed between each stage. Specifically, the global attention query $\mathbf{G}_q$ starts with a matrix of size $B\times C \times h \times w$, where $B$, $C$, $h \times w$ refers to batch size, channels, and spatial dimensional of the local window. In this way, the global query generator will repeat along batch dimension, and then be reshaped and added into multiple heads of local self-attention modules.

\textbf{Global Self-Attention:} Based on the global query token, the global self-attention can be formulated as follows:

\begin{equation}
\text{Global\_Attention}(\mathbf{g}_q, \mathbf{k}, \mathbf{v}) = \text{Softmax}( \frac{\mathbf{g}_q \mathbf{k}}{\sqrt{s}} + \mathbf{p}) \mathbf{v}
\end{equation}

where $s$, $\mathbf{p}$ refers to a scaling factor and a learnable relative position embedding vector. For instance, if the image patch position ranges from $[-b + 1, b - 1]$ then $\mathbf{p}$ will be generated based on spatial positions 
from a spatial grid of $\sR^{(2b-1)\times (2b-1)}$. In this way, local self-attention has access to even long-range information from imagery regions outside of local windows, which provides an effective way of extending the reception field of self-attention without increasing the computation complexity. 

In this paper, the CGCViT is able to extend this state-of-the-art ViT model into a dual branch setting and provide a rock-solid backbone for the cross-view damage classification task.


%% file: experiment.tex
\section{Experiment} \label{experiment}

\subsection{Dataset overview} \label{data_overview}

\begin{figure*}[!htbp]
\centering
\centering
\includegraphics[width=\linewidth]{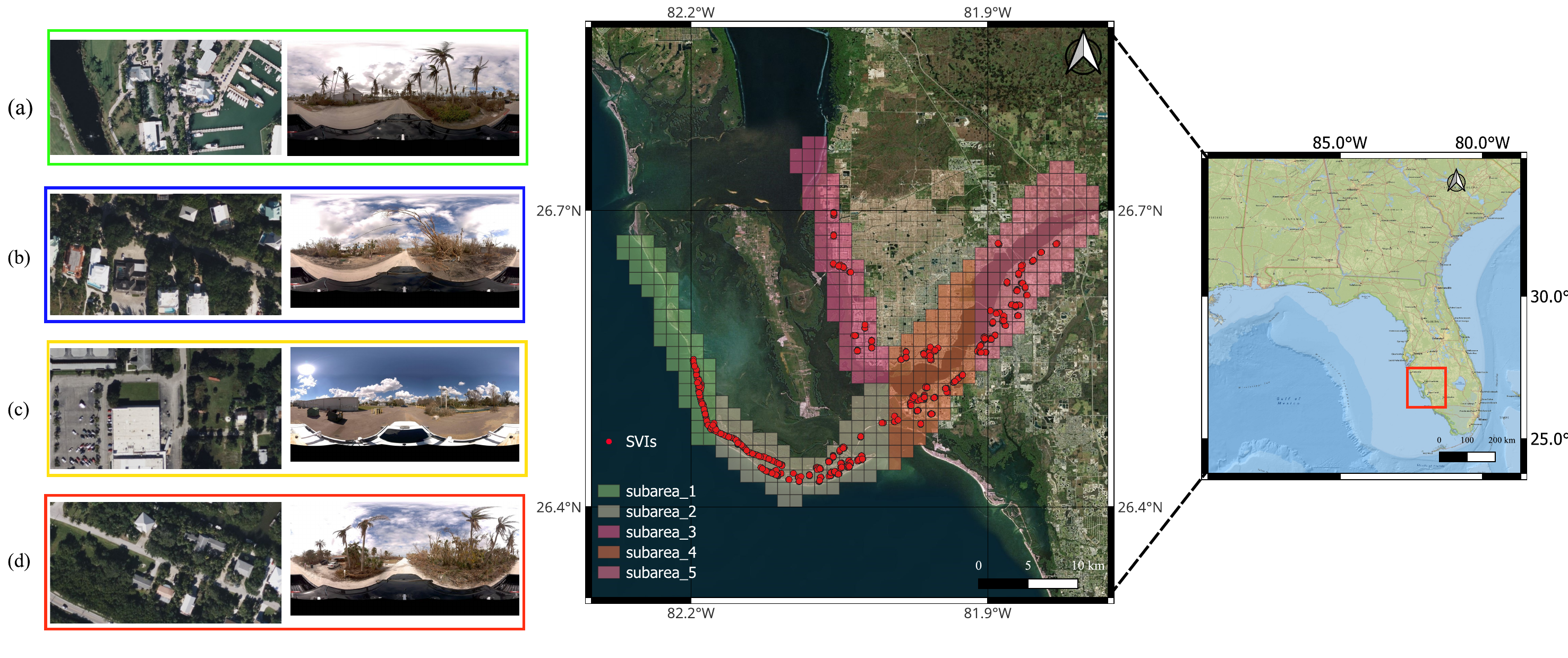}
\caption{Overview of the study area together with the street-view and VHR satellite imagery. Five subareas are depicted in different colors in the middle image, where (a) to (d) are selected cross-view imagery pairs of \model{}. }
\label{fig:dataset}
\end{figure*}

\begin{table*}[!htbp]
\caption{Overview of the \dataset{} dataset, split into  five subareas, respectively.}
\label{tab:dataset_statics}
\centering
\begin{adjustbox}{width=0.8\linewidth}
\begin{tabular}{@{}lccccc@{}}
\toprule
\toprule
\textbf{Subarea} & \textbf{VHR Reso (cm)} & \textbf{Image Pixel (Rows, Columns)} & \textbf{Num of raw SVIs} & \textbf{Num of selected SVIs}  & \textbf{Aera (km2)} \\ \midrule
1  & 27 & 37,137 $\times$ 78,833 & 7,511 & 112 & 80.53 \\
2    & 27 & 64,934 $\times$ 46,403 & 42,987 &386 & 134.40 \\
3     & 27 & 37,137 $\times$ 92,732 & 59,932 &112 & 133.96 \\ 
4     & 27 & 37,137 $\times$ 78,833 & 158,605 &271 & 125.70 \\ 
5     & 27 & 46,403 $\times$ 78,833 & 688,504 &254 & 177.77 \\ 
\bottomrule
\bottomrule
\end{tabular}
\end{adjustbox}
\end{table*}

\begin{figure*}[!htbp]
\centering
\centering
\includegraphics[width=0.9\linewidth]{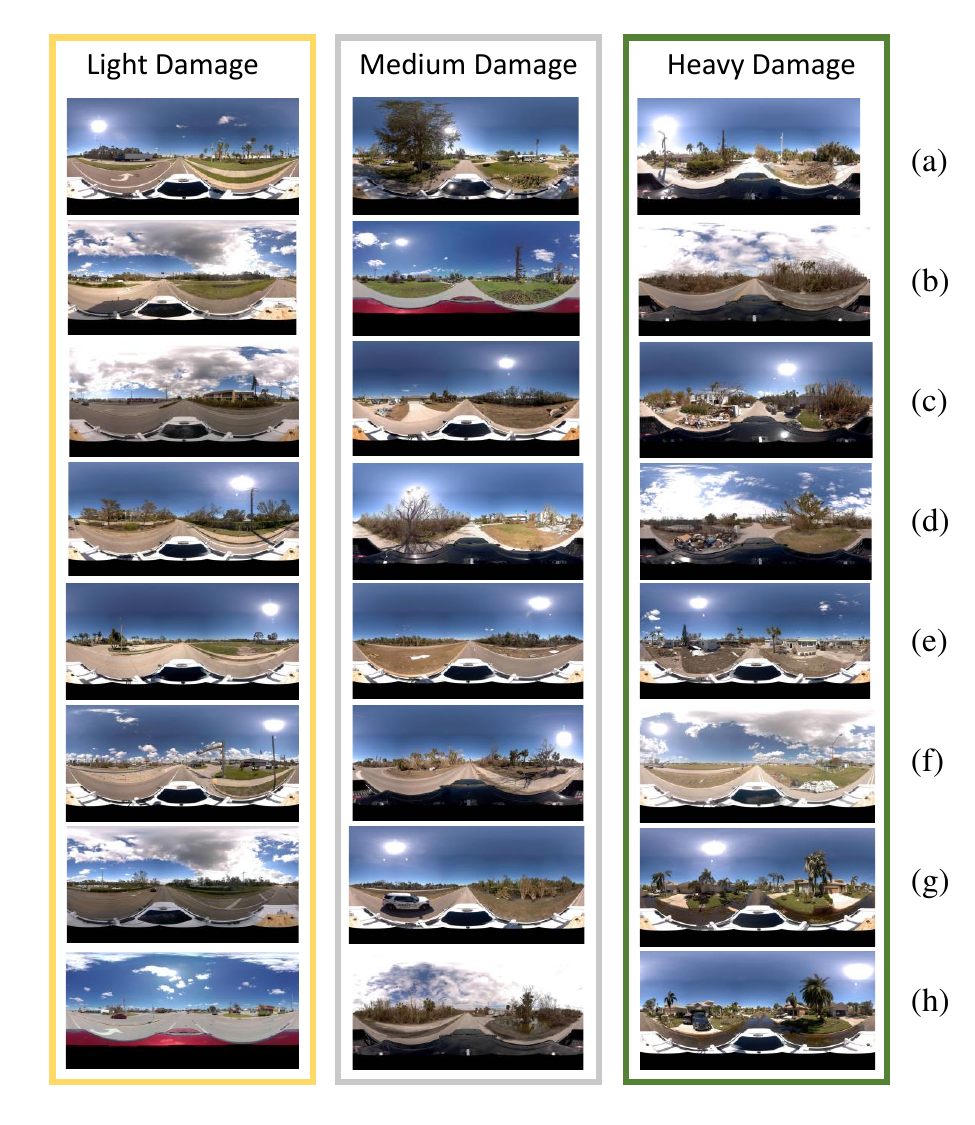}
\caption{Stree-view imagery-based damage perception classification criteria with three level damages (light, medium, and heavy damage from low to high). (a)-(b): damage perception estimated based on fallen trees, (c)-(d): damage perception estimated based on housing trash, (e)-(f): damage perception estimated based on street signs or destroyed buildings, (g)-(h): damage perception estimated based on standing water in the street. 
}
\label{fig:SVI_data}
\end{figure*}

\begin{figure*}[!htbp]
\centering
\centering
\includegraphics[width=\linewidth]{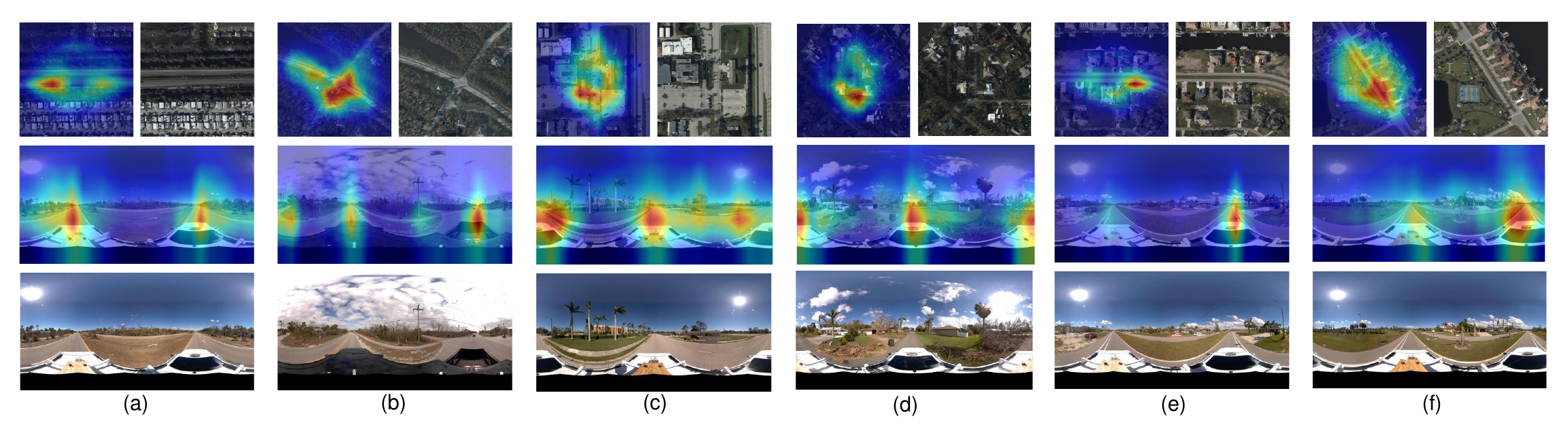}
\caption{Heatmaps for correct geolocalized imagery pairs together with raw cross-view SVI and satellite imagery.}
\label{fig:geoloca_results}
\end{figure*}

Hurricane IAN formed on September 23, 2022, causing severe storm surges and significant economic losses, making it one of the most devastating hurricanes in the history of Florida, USA. To this end, we have selected the renowned Sanibel Island and its surrounding area in southwest Florida, which was hit devastatingly by Hurricane IAN in 2022, as our case study area and created a novel cross-view dataset, namely \dataset{}.

\textbf{VHR Satellite Imagery:}
VHR satellite imagery provides extremely detailed overhead surface information, which is crucial for assessing disaster impacts, planning rescue operations, and formulating recovery strategies.
For Hurricane IAN, the National Oceanic and Atmospheric Administration (NOAA) has collected relevant VHR satellite imagery. Each image is assembled into a mosaic distributed in tiles, with a ground sample distance of approximately 15 to 30 cm per pixel. In this study, we selected VHR imagery from September 30, 2022 from the NOAA open data portal\footnote{https://storms.ngs.noaa.gov/storms/ian}, and divided it into five subareas to support the assessment of Hurricane IAN's damage extent. These images provide a fine-grained head-view of the study area right after the hurricane, thus enhance our understanding of the impact of the disaster and aid in developing effective response and recovery measures.

\textbf{Street-view Imagery :} The street-view images of Hurricane IAN used in our study were collected from the open-source Mapilliry platform, specifically from a mapping campaign conducted by Site Tour 360 in our study area. These images were captured by Site Tour 360 after access was restored post-disaster. Site Tour 360 utilized Mapillary as an mapping tool. The Mapillary platform can rapidly and openly disseminate these high-resolution images, which is crucial for disaster response. This enables rescue organizations and the public to promptly access the latest post-disaster images, aiding in identifying areas in urgent need of assistance and efficiently allocating resources.

For downloading and filtering these street-view images, we adopted the ZenSVI \footnote{https://github.com/koito19960406/ZenSVI} tool. ZenSVI can efficiently download, process, and analyze large-scale street-view image data, providing valuable data support for planning post-disaster recovery efforts. In total, we have processed and filtered in total 957,539 SVI records from Mapiliary using geographic extents (i.e., five subareas) and their timestamps (i.e., only after 28th September, 2022), out of which we have selected 1,135 and manually labelled them for the damage perception level with a group of GIS and disaster experts. The detailed split of SVI and extent of VHR satellite imagery is listed in Table \ref{tab:dataset_statics}.

\textbf{Damage Perception Reference Data:} Based on the aforementioned 1,135 SVI related to Hurricane IAN, we manually categorized them into three damage severity levels - light, medium, and heavy damages - based on a list of quantifiable and disaster-related indicators. Specifically, light damage images are characterized by a clean scene with no significant damage or only light damage, such as small areas of fallen trees or a few small road signs knocked down. Medium damage images are relatively cluttered and typically include larger or more extensive areas of fallen trees, as well as standing water around the trees. These images may also show more fallen road signs or road closure signs. Heavy damage images are very chaotic, featuring large or extensive areas of fallen trees, flooded roads, and housing trash. These indicators provide a extensible and subjective basis of disaster damage perception, which serves as the reference data for the subsequent cross-view imagery classification and validation.

As shown in Figure \ref{fig:SVI_data}, we have elaborated some exemplary SVI in our \dataset{} dataset with light, medium, and heavy damage based on different damage indicators. Among them, (a) and (b) are classified based on the amount fallen trees. (c) and (d) are classified according to the amount of housing trash. (e) and (f) are classified based on the damage to road signs or destroyed buildings. (g) and (h) are classified according to the extent of standing water. This completes the damage perception reference data.

\subsection{Experiment setup for Cross-View Geolocalization}
In our experimental setup for \geolocmodel{}, we employed a ConvNeXt-Base model, initialized using a pre-trained Sample4Geo model~\citep{deuser2023sample4geo}. This pre-training on CVUSA allowed us to leverage a robust feature extraction as CVUSA features rural and urban environments. During pre-processing, we made sure that the street-view images were oriented north according to the CVUSA standard. We also cropped the top and bottom of the images to reduce their size and eliminate irrelevant information, thereby streamlining the input data for more efficient processing. As a result, the cross-view imagery pairs are of size $512\times1024$ pixels and $512\times512$ pixels for SVI and satellite imagery, respectively. During training, we used the InfoNCE loss function with label smoothing set to 0.1. This regularization technique helped to mitigate overconfidence in the predictions, thus promoting better generalization.

The model fine-tuning process was performed over 10 epochs using the AdamW \citep{kingma2014adam} optimizer with an initial learning rate of $0.0001$. We use a learning rate scheduler with a warm-up of one epoch and a cosine decay for the remaining epochs. To address potential overfitting and improve generalization, we incorporate several data augmentation techniques. For both images, we use synchronous horizontal flipping and rotation to ensure consistent orientation with the corresponding street view images. In addition, we applied grid dropout and coarse dropout to prevent the model from focusing excessively on certain regions of the images. Color jittering is also used to improve the model's robustness to variations in lighting conditions.

As for \imgclsmodel{}, we implemented the CGCViT model based on a backbone network of GCViT-Tiny with 20 million parameter pre-trained on ImageNet-1K dataset \citep{deng2009imagenet}. Herein, we used exact the same cross-view imagery pairs as in \geolocmodel{} with $512\times1024$ pixels for SVI and $512\times512$ pixels satellite imagery as a input feature size. Specifically, we fine-tuned the CGCViT with the AdamW \citep{kingma2014adam} optimizer for 100 epochs with an initial learning rate of 0.03, weight decay of 0.05 with acosine decay scheduler, and 10 warm-up epochs, respectively.

\begin{table*}[!htbp]
\caption{Performance Metrics for Different Pre-trained Geolocalization Models}
\label{tab:pretrained_metrics}
\centering
\begin{tabular}{@{}lccccc@{}}
\toprule
\toprule
Method & Fine-tuned & R@1 & R@5 & R@10 & R@1\% \\ \midrule
TransGeo ~\cite{zhu2022transgeo}&  \xmark  & 38.30 & 67.99 & 79.34 & 97.25 \\
SAIG-D  ~\cite{zhu2023simple}&   \xmark & 43.62 & 72.43 & 83.33 & 98.05 \\ 
Sample4Geo ~\cite{deuser2023sample4geo} & \xmark & 74.56 & 91.22 & 95.48 & \textbf{99.11} \\ \midrule
\geolocmodel{} (2:8)  & \cmark &\textbf{81.84} & \textbf{96.68} & \textbf{98.45} & 98.34\\
\bottomrule
\bottomrule
\end{tabular}
\end{table*}

\begin{table*}[!htbp]
\caption{Performance Metrics for Fine-tuned Geolocalization Models with Different Train/Test Ratios.}
\label{tab:finetuned_metrics}
\centering
\begin{tabular}{@{}lcccccccccccccc@{}}
        \toprule
        \toprule
        train:test & \multicolumn{2}{c}{1:9} & \multicolumn{2}{c}{2:8} & \multicolumn{2}{c}{3:7} & \multicolumn{2}{c}{4:6} & \multicolumn{2}{c}{5:5} & \multicolumn{2}{c}{6:4} \\
        \cmidrule(lr){2-3} \cmidrule(lr){4-5} \cmidrule(lr){6-7} \cmidrule(lr){8-9} \cmidrule(lr){10-11} \cmidrule(lr){12-13}
        Fine-tuned & \xmark & \cmark & \xmark & \cmark & \xmark & \cmark & \xmark & \cmark & \xmark & \cmark & \xmark & \cmark \\
        \midrule
        Recall@1 & 75.69 & 79.72 & 75.97 & 81.84 & 78.86 & 87.47 & 79.03 & 87.74 & 83.33 & 92.20 & 82.30 & 91.37 \\
        Recall@5 & 92.22 & 95.87 & 92.47 & 96.68 & 93.80 & 97.97 & 94.39 & 98.23 & 96.99 & 99.47 & 96.02 & 98.89 \\
        Recall@10 & 96.06 & 98.13 & 96.23 & 98.45 & 97.47 & 99.11 & 97.49 & 99.26 & 98.94 & 99.82 & 98.67 & 99.56 \\
        Recall@top1 & 96.06 & 98.13 & 96.01 & 98.34 & 95.57 & 98.99 & 94.98 & 98.52 & 96.99 & 99.47 & 95.13 & 98.89 \\
        \bottomrule
        \bottomrule
\end{tabular}
\end{table*}

\begin{table*}[!htbp]
    \centering
    \caption{Class-wise Performance Comparison of Precision (P), Recall (R) and F1 score for SVI, VHR Satellite, and \imgclsmodel{}.}
    \begin{tabular}{lcccccccccc}
        \toprule
        \toprule
        & \multicolumn{3}{c}{SVI} & \multicolumn{3}{c}{VHR Satellite} & \multicolumn{3}{c}{\imgclsmodel{}} \\
        \cmidrule(lr){2-4} \cmidrule(lr){5-7} \cmidrule(lr){8-10}
        & P & R & F1 & P & R & F1 & P & R & F1\\
        \midrule
        Light & 78.61 & \textbf{88.15} & 0.83 & 72.52 & 82.37 & 0.77 & \textbf{82.64} & 87.64 &  \textbf{0.85}\\
        Medium & 62.04 & 56.30 & 0.59 & 55.69 & 39.08 & 0.45 & \textbf{64.29} & \textbf{66.81} & \textbf{0.65}\\
        Heavy & 81.88 & 72.19 & 0.76 & 65.80 & \textbf{75.15} & 0.70& \textbf{89.15} & 71.88 &  \textbf{0.79}\\
        \bottomrule
        \bottomrule
    \end{tabular}
    \label{tab:classification_metrics}
\end{table*}

\begin{table}[!htbp]
    
    \caption{Overall Performance Metrics for SVI, Satellite, and \imgclsmodel{}.}
    \label{tab:classification_overall}
    \centering
    \begin{tabular}{lcccc}
        \toprule
        \toprule
        & P & R & OA & F1\\
        \midrule
        SVI & 74.17 & 72.21 & 74.50 & 0.73\\
        VHR Satellite & 64.67 & 65.69 & 67.07 & 0.65 \\
        \imgclsmodel{}(5:5) & \textbf{78.69} & \textbf{75.44} & \textbf{77.96} & \textbf{0.77}\\
        \bottomrule
        \bottomrule
    \end{tabular}
\end{table}

\subsection{Geo-localization results}

In our evaluation, we start with comparing state-of-the-art cross-view geolocalization models that were pre-trained on the CVUSA data, specifically TransGeo ~\citep{zhu2022transgeo}, SAIG-D  ~\citep{zhu2023simple}, and Sample4Geo ~\citep{deuser2023sample4geo}. First, we directly apply three pre-trained models on the \dataset{} dataset to serve as a baseline of cross-view geolocalization performance. Next, we compare the fine-tuned model against pre-trained baseline models and conduct an ablation study w.r.t the ratio of train and test samples, ranging from 1:9 to 6:4. Since the \geolocmodel{} is formulated as an imagery retrieval task, we consider four Recall@K evaluation metrics, namely, Recall@1, Recall@5, Recall@10, and Recall@1\%, where K refers to the top K imagery that given by the query. A higher value (i.e., ranging from 0 to 100) simply means better accuracy.

\textbf{Pre-trained Cross-view Geolocalization}: Table \ref{tab:pretrained_metrics} show a few interesting findings: 
1) all three pre-trained models form a nice baseline of addressing cross-view geolocalization tasks in a completely unseen area with a R@10 around 80\%. Noticeably, the pre-training dataset, namely the CVUSA dataset, differs with the \dataset{} to a large extend in both SVI (one from GSVI and one from Mapillary) as well as Satellite imagery. These results confirm the promising value of cross-view geolocalization approaches as a pure vision-based alternative to classic positioning techniques (e.g., GPS, Wifi), especially in a disaster response scenario; 
2) Given 30\% of \dataset{} imagery pairs for fine-tuning, \geolocmodel{} achieves significant performance boosting at almost all Recall@K metrics (i.e., except Recall@1\%) against three baseline models, with a relatively small improvement compared to Sample4Geo as they shared similar network architectures. This means one can easily adapt pre-trained cross-view geolocalization models to a new case study area with a limited cost of preparing "warming-up" contrastive learning samples for a much more affordable fine-tuning process than training an entirely new model from scratch, which will be an important feature desired for timely disaster response and geolocalization usage. In short, the preliminary results from implementing \geolocmodel{} model on Hurricane IAN uncovers a promising avenue for leveraging pre-trained cross-view geolocalization techniques for low-cost and weather-resilient location awareness with such a time-critical task.

Moreover, by visualizing the correct geolocalized imagery pairs in Figure \ref{fig:geoloca_results} (a) to (f), we see that the \geolocmodel{} model was able to correlate SVI and satellite imagery based on landmarks, such as street, crossroad, building, which is similar to how human will spatially geolocalize themselves in an unknown place. Unlike to existing benchmarks, the \dataset{}dataset is featured by massive and diverse damages (e.g., for streets, buildings, and vegetation as shown in Figure \ref{fig:SVI_data}) caused by Hurricane IAN, which pose a unique challenge for our \geolocmodel{} model. Unsurprisingly, the heatmap visualization confirms the robustness of our model in handling such sophisticated structure changes in the built environment at a much affordable cost and additional effort. From our perspective, these results can motivate future works to investigate even spatial-temporal changes in a cross-view setup.

\textbf{Ablation study}: As an ablation study, we examine the effect of different train and test ratios on the model performance of \geolocmodel{}. Specifically, we consider two variables here: 1) pre-trained or fine-tuned models, 2) how many samples are used for fine-tuning. Table \ref{tab:finetuned_metrics} shows the comparative performance changes w.r.t these two variables. Two key findings deserve extra attention: 
first, an obvious finding is that more fine-tuning samples lead to generally higher performance boosting with an exception from 5:5 to 6:4, where the fine-tuned model performance start to drop. A potential reason is related to the size of our \dataset{} dataset and its relatively small geographical coverage, which can be a future work direction to consolidate the finding here; second, we see already a satisfying performance boosting using limited fine-tuning with up to a few hundred imagery pairs (e.g, 1:9 and 2:8). This provides extra flexibility and reduced deploying time for the proposed framework during a real-world disaster response.

\subsection{Disaster mapping results}

\begin{figure*}[!htbp]
\centering
\centering
\includegraphics[width=\linewidth]{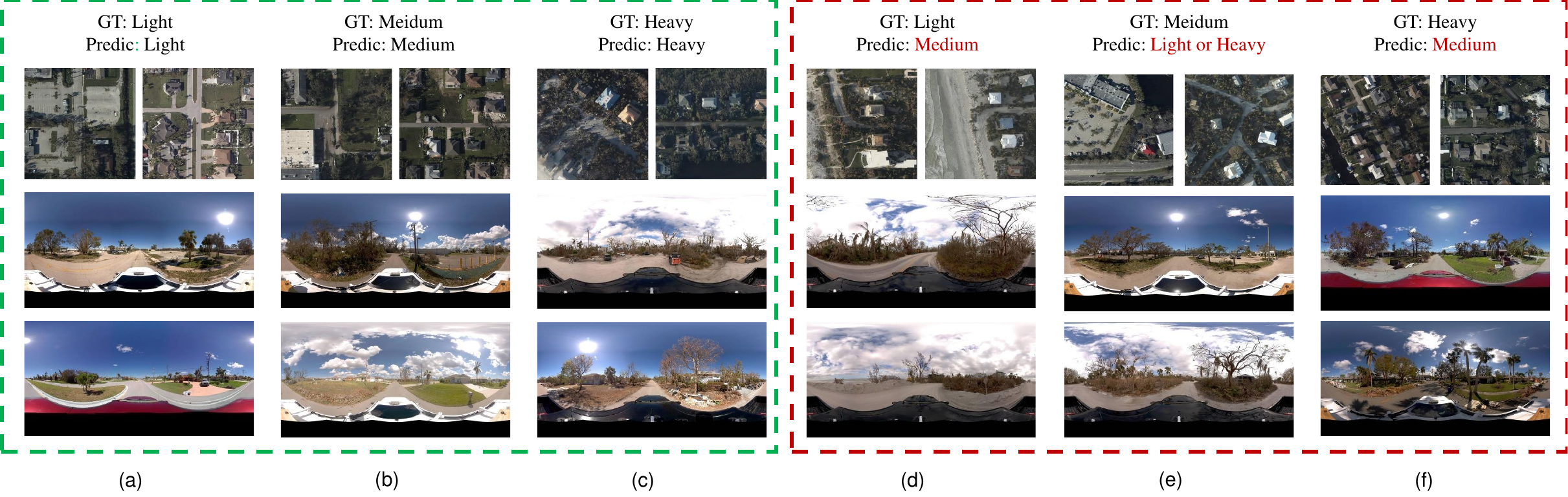}
\caption{Visualizations of \imgclsmodel{} classification results. 
(a) to (c) are correctly predicted imagery pairs and (d) to (f) are wrongly predicted imagery pairs. In each image, we show both Ground Truth (GT) and predicted (Predic) labels.}
\label{fig:classification_results}
\end{figure*}

To evaluate the \imgclsmodel{} model on cross-view damage perception estimation, we first compare the CGViT model trained on cross-view imagery pairs with two single-view models (i.e., one trained on SVI and one trained on VHR satellite imagery). Next, we conduct a similar ablation study as in \geolocmodel{} to investigate the effect of fine-tuning ratios on the model performance. In this context, we consider mainly four multi-classification metrics, namely Precision (P), Recall (R), Overall Accuracy (OA), and the F1 score.

\textbf{Cross-view Damage Perception Estimation}: Table \ref{tab:classification_metrics} and \ref{tab:classification_overall} compare the performance of our \imgclsmodel{} model with two single-view models based on either SVI or VHR satellite imagery. Specifically, Table \ref{tab:classification_metrics} shows a detailed class-wise evaluation metrics w.r.t three level of damage labelled in the \dataset{} dataset (i.e., Light, Medium, and Heavy Damages). A clear pattern is that medium damages are often more challenging (with low F1 scores in all three cases) than light and heavy damages. This can be attributed to how the damage perception levels are classified in the reference dataset (see Figure \ref{fig:dataset}) as medium damages involves both qualitative and quantitative analysis of those damage indicator we considered (as introduced in Section \ref{data_overview}), thus pose a general challenge to disaster mapping approaches \citep{dong2013comprehensive}. More importantly, we see a stimulating accuracy improvement when extending single-view to cross-view, for instance, \imgclsmodel{} outperforms both SVI and VHR satellite in all three classes in P and F1 score. Nevertheless, Table \ref{tab:classification_overall} confirms the advantage of using cross-view imagery for disaster perception estimation rather than any single-view models.

\begin{table}[!htbp]
    \centering
    \caption{Performance Metrics for Cross-View Damage Perception Estimation with different Train Test ratios.}
    \begin{tabular}{lcccccc}
        \toprule
        \toprule
        train:test & 1:9 & 2:8 & 3:7 & 4:6 & 5:5 & 6:4 \\
        \midrule
        
        P & 65.65 & 73.00 & 68.87 & 73.63 & 78.69 & 70.99 \\
        R & 65.36 & 72.12 & 69.06 & 72.45 & 75.44 & 70.36 \\
        OA & 66.84 & 73.80 & 70.05 & 73.89 & 77.96 & 71.31 \\
        F1 &  0.66 & 0.73 & 0.69 & 0.73 & 0.77 & 0.71 \\        
        \bottomrule
        \bottomrule
    \end{tabular}
    \label{tab:classification_ablation}
\end{table}

Moreover, the aforementioned statement can be strengthened when we start to visualize the \imgclsmodel{} classification results w.r.t where it works and where it fails. Figure \ref{fig:classification_results} demonstrates both correct and incorrect classification cases using the cross-view \imgclsmodel{} model. The correct cases (e.g., Figure \ref{fig:classification_results}(a) to (c)) basically echo the previous finding where SVI provides major hints based on various damage indicators (e.g., fallen trees, housing trash, etc). As for incorrect examples (e.g., Figure \ref{fig:classification_results} (d) to (f)), one interesting pattern is that though heavy damage might be misclassified as medium damage, there is not a single case that light damage cases are classified as heavy ones. Herein, medium damages remain challenging as they may be confused with both light and heavy damages.

\textbf{Ablation Study}: Similarly to \geolocmodel{}, we conduct an ablation study to examine the influence of train and test ratios in the classification performance as listed in Table \ref{tab:classification_ablation}. Herein, one can see that the training process of cross-view damage perception estimation involves more uncertainties when exposed to an increasing number of training samples. Our assumption is that disaster damages are often spatial auto-correlated due to environmental and human factors, thus a random sampling is insufficient to ensure all possible damage types are well-covered. In this context, a possible solution is to explicitly consider the spatial auto-correlation of cross-view imagery early enough in the sampling process, such as using metric auto-correlation \citep{wang2024mc} or locality sensitive sampling \citep{luo2019scaling}. 

\begin{figure}[!htbp]
\centering
\centering
\includegraphics[width=\linewidth]{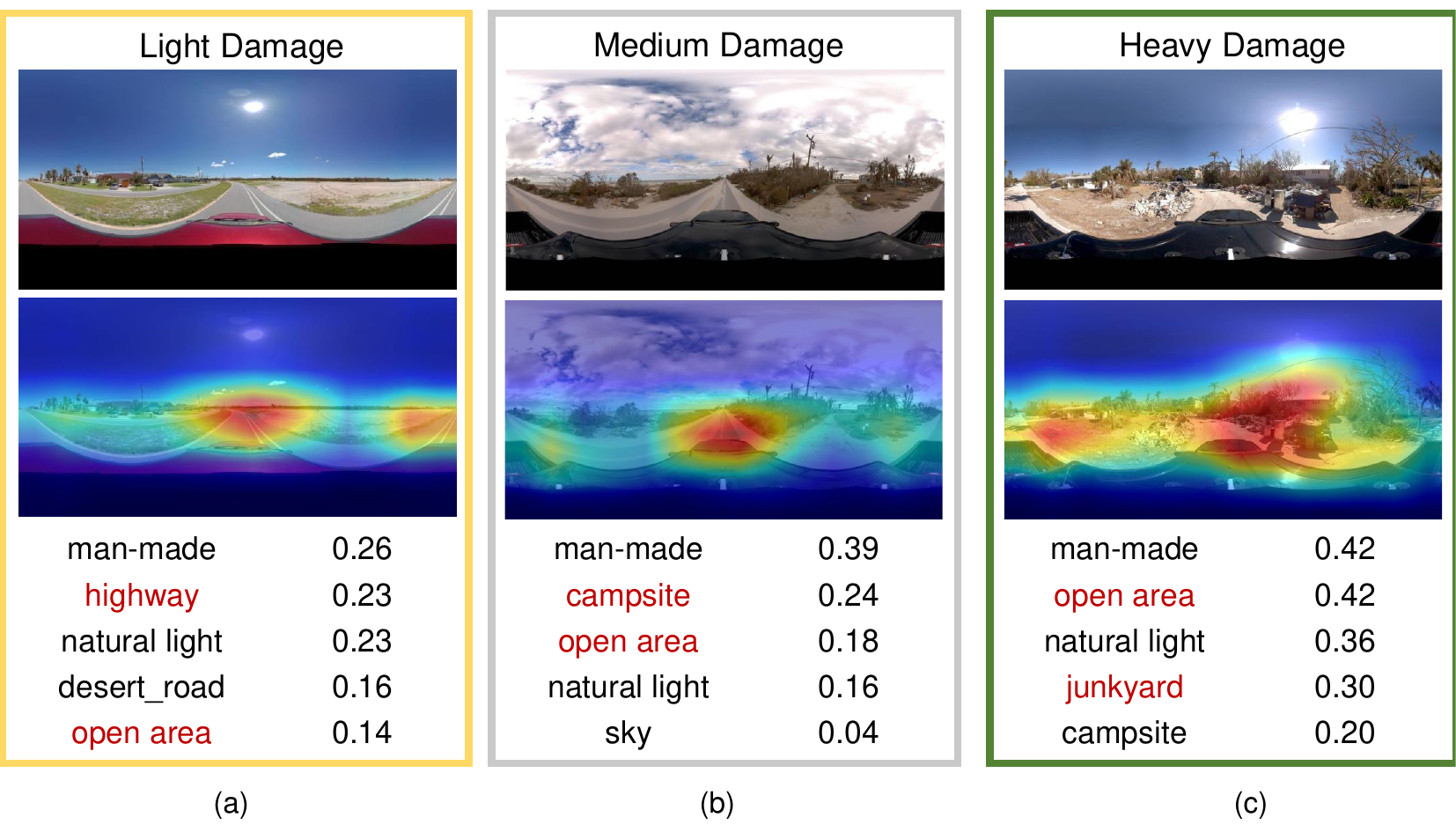}
\caption{Selected example of SVI scene classification with Place365 classes for three-level damages in the \dataset{} dataset (from left to right: light, medium, and heavy damage).
}
\label{fig:SVI_scene_classification}
\end{figure}

%% file: discussion.tex
\section{Discussions} \label{dicussion}

\begin{figure*}[!htbp]
\centering
\centering
\includegraphics[width=\linewidth]{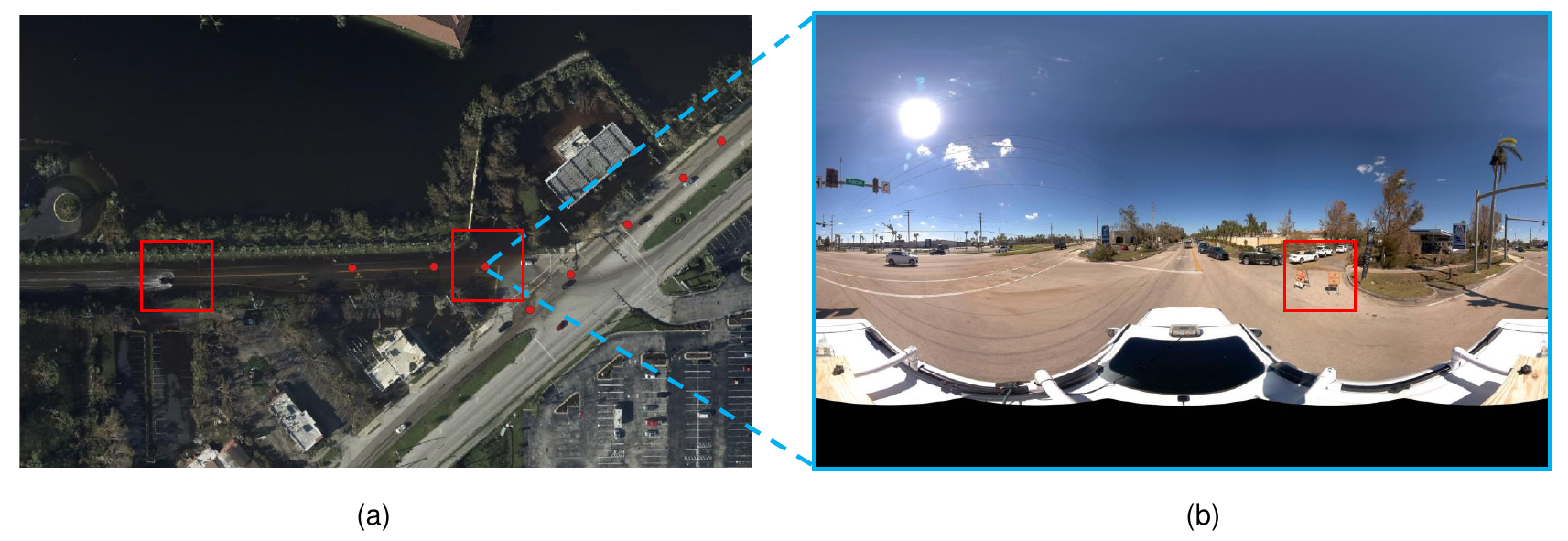}
\caption{A Challenge case in the CVDisaster dataset. (a) VHR satellite imagery from 30th September, which is right after Hurricane IAN landed in the city; (b) SVI in Mapillary collected on 2nd October, where water on the street has already been cleaned but a road-sign of "Detour" was left as a proof of damage caused by Hurricane IAN. }
\label{fig:challenge_case}
\end{figure*}

In this paper, we propose \model{}, as a novel framework to simultaneously tackle two important tasks in a disaster response scenario, which are cross-view geolocalization and disaster perception estimation. As a case study, we constructed a first-of-this-kind dataset (i.e., the \dataset{} dataset) based on SVI and VHR satellite imagery collected around Sanibel Island after Hurricane IAN, based on which we conducted extensive evaluation of the proposed framework and its major components. Despite the promising results, there are a few limitations that deserve future attention: 1) Though pre-trained geolocalization models offer a good baseline performance for \geolocmodel{}, \model{} still requires training efforts mainly to learn and align the classification feature space (\imgclsmodel{}) to various damage indicators spotted by human experts. Therefore, a future work direction is to automatically extract and quantify those damage related indicators from SVI, for instance by adopting the Place365 scene classification categories (Figure \ref{fig:SVI_scene_classification}). Of course, it would be great if disaster-related scenes or targets, such as flooding or housing trash, can be added to those pre-training datasets. To this end, we see \model{} make a unique contribution towards inspiring a list of integrated cross-view applications with the GeoAI research community. 
2) there are still many unsolved challenges in such a cross-view disaster mapping scenario. For example, Figure \ref{fig:challenge_case} shows an interesting case where from the VHR satellite imagery that NOAA collected on 30th September 2022 (two days after Hurricane IAN), one can observe massive standing water on the street and cares need to risk for access. However, from the Mapillary SVI taken on 2nd October (four days after Hurricane IAN), water is cleaned with only road signs ("Detour") left as a sign of potential damages. Together with the inherent ambiguity of the damage perception level, the rapid temporal change during the disaster poses another level of difficulty to this task. Future cross-view models should definitely take such temporal changes into considering during the model pre-training. Last but not the least, our future work will focus on extending the \dataset{} dataset to cover multiple areas and disaster types around the world.

%% file: conclusion.tex
\section{Conclusions} \label{conclusion}

In this work, we present a novel framework, namely \model{}, to a time-crucial application scenario of disaster mapping, where two types of information are key, which are disaster damage perception and geolocation awareness. To the best of our knowledge, \model{} is the first of this kind framework that can simultaneously achieve cross-view geolocalization (\geolocmodel{}) and disaster damage perception estimation (\imgclsmodel{}). A case study on the \dataset{} dataset collected from Hurricane IAN confirms the advantages of our \model{} framework over classic positioning techniques (e.g.., GPS, Wifi) as well as damage assessment approaches purely based on Very High Resolution (VHR) satellite imagery. We show that one can achieve highly competitive performance (over 80\% for geolocalization and 75\% for damage perception estimation) with limited fine-tuning efforts by benefiting from state-of-the-art pre-trained vision models, like ConvNeXt and CGCViT. We hope our findings can motivate future cross-view models and applications within a broader GeoAI research community.